\documentclass{bmvc2k}

\title{How to Manipulate CNNs to Make Them Lie: the GradCAM Case}

\addauthor{Tom Viering}{t.j.viering@tudelft.nl}{1}
\addauthor{Ziqi Wang}{z.wang-8@tudelft.nl}{1}
\addauthor{Marco Loog}{m.loog@tudelft.nl}{12}
\addauthor{Elmar Eisemann}{e.eisemann@tudelft.nl}{1}

\addinstitution{
 Delft University of Technology\\
 Delft, The Netherlands
}
\addinstitution{
 University of Copenhagen\\
 Copenhagen, Denmark
}

\runninghead{Viering, Wang, Loog, Eisemann}{How to Make CNNs Lie: the GradCAM Case}

\usepackage{url}
\usepackage{multirow}
\usepackage{graphicx}
\usepackage{amsmath}
\usepackage{booktabs}
\usepackage{algorithm}
\usepackage{algorithmic}
\usepackage{bbold}

\usepackage{latexsym} 
\usepackage{placeins}

\begin{document}

\maketitle

\begin{abstract}
Recently many methods have been introduced to explain CNN decisions. However, it has been shown that some methods can be sensitive to manipulation of the input. We continue this line of work and investigate the explanation method GradCAM. Instead of manipulating the input, we consider an adversary that manipulates the model itself to attack the explanation. By changing weights and architecture, we demonstrate that it is possible to generate any desired explanation, while leaving the model's accuracy essentially unchanged. This illustrates that GradCAM cannot explain the decision of every CNN and provides a proof of concept showing that it is possible to obfuscate the inner workings of a CNN. Finally, we combine input and model manipulation. To this end we put a backdoor in the network: the explanation is correct unless there is a specific pattern present in the input, which triggers a malicious explanation. Our work raises new security concerns, especially in settings where explanations of models may be used to make decisions, such as in the medical domain.
  \end{abstract}

\section{Introduction}

For deep convolutional neural networks, it is difficult to explain how models make certain predictions. Explanations for decisions of such complex models are desirable \citep{surveyXAI}. For example, in job application matching, explanations may reveal undesireable biases in machine learning models. For settings which demand rigorous security demands such as self driving cars, explanations can help us better understand how models work in order to identify and fix vulnerabilities. In other application domains, such as neuroscience, machine learning is not only used for predictions (e.g., regarding a disease), but also to understand the cause (the underlying biological mechanism). In this case, explanations can help domain experts discover new phenomena.

The field of Explainable AI (XAI) aims to tackle this problem; how did a particular model come to its prediction? For CNNs a popular explanation takes the form of heatmaps or saliency maps \citep{simplegrad}, which indicate the pixels that were important for the final output of the model. Recently, many explanation techniques have been proposed in the literature to generate explanations for machine learning models~\citep{SHAP,GradCAM,patternnet,AxiomaticAttribution,simplegrad,guidedbackprop,deconvnet_original,koh2017understanding,gradientTimesInput,bach2015pixel,smoothgrad,LIME,amogh_localization,Fong2017InterpretablePerturbation,Dabkowski2017RealClassifiers,Zintgraf2017VisualizingAnalysis,objectdetector}. A nice introduction and survey to the XAI is~\citep{surveyXAI}.

Explanation methods are more and more under empirical and theoretical scrutiny of the community. For example, \citet{unifiedGradients} show equivalence and connections between several explanation methods, and \citet{SHAP} unify six existing explanation methods. Several studies  \citep{AxiomaticAttribution,patternnet,sanity,unreliable,fragile} 
have raised questions regarding robustness and faithfulness of these explanations methods. For example, \citet{fragile} show that an adverserial imperceptible perturbations of the input can change the explanation significantly while the model's prediction is unchanged. 

We continue this line of investigation and uncover new (security) vulnerabilities in the popular explanation method GradCAM \citep{GradCAM}. GradCAM, a generalization of the explanation method CAM \citep{Zhou_2016_CVPR}, is a fast and simple method to explain CNN decisions and is applicable to many CNN architectures. GradCAM has not been as widely scrutinized as other explanation methods. \citet{sanity} propose several sanity checks that should be satisfied by explanation methods, e.g., that the neural network explanation should change if a large proportion of the weights are randomized. \citet{sanity} find GradCAM satisfies their proposed checks, motivating further study of this explanation method.  

Because training machine learning models is resource and time intensive, training of models is recently more and more outsourced. It is now possible to upload training data and model architecture, and to train the model in the cloud, for example using platforms created by Google \citep{cloud_google}, Amazon \citep{cloud_amazon} or Microsoft \citep{cloud_azure}. It is expected that this will become the norm. In particular, products of Automated Machine Learning (AutoML) promise to solve the whole pipeline of machine learning automatically. The user only has to upload the dataset, and the cloud provider will automatically try several architectures, tune hyperparameters, train models, and evaluate them \citep{automl_survey}.
Another approach to circumvent costly training procedures is to finetune existing models for new tasks \citep{transferlearning}. 

Both outsourcing and finetuning pose a security risk \citep{badnets}. \citet{badnets} show in their case study with traffic signs, that by manipulating the training data, the model will misclassify stop signs if a sticker is applied to them. \citet{trojan} introduce a technique that can be applied to an already trained model to introduce malicious behaviour. Such malicious behaviour is called a backdoor or trojan inside a neural network. The backdoor is triggered by specific input patterns while keeping model performance on the original task more or less the same. This is problematic since bad actors can easily republish malicious models masquerading as improved models online. Because of the blackbox nature of deep learning models, such trojans are difficult to detect \citep{cleanse,Chen2019DetectingClustering}. Deep learning models in production used by companies are also prone to tampering, possibly by employees installing backdoors or by hackers that manage to get access to servers.

In this work, instead of examining robustness of explanations with respect to a changing input as investigated by \citet{fragile}, we investigate the robustness of explanations when the model is modified by an adversary such as the scenario considered by \citet{trojan} and \citet{cleanse}. Our work can be considered as a white-box attack on the explanation method GradCAM and the model \citep{Papernot2016TowardsLearning}.

\begin{figure}[htb]
\centering

\subfigure[Input image]{
\begin{minipage}{0.3 \textwidth}
\centering
\includegraphics[width = 1 \textwidth]{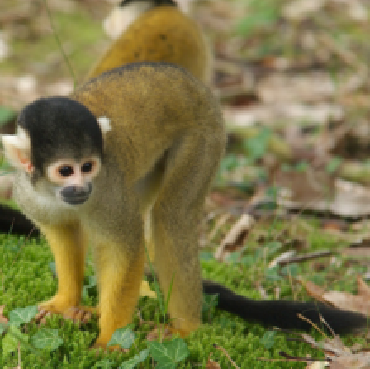}
\label{fig_input}
\end{minipage}
}
\subfigure[Explanation of original CNN]{
\begin{minipage}{0.3 \textwidth}
\centering
\includegraphics[width = 1 \textwidth] {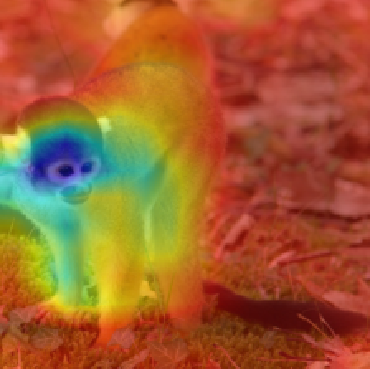}
\label{fig_expl_GT}
\end{minipage}
}
\subfigure[Expl. T1]{
\begin{minipage}{0.3 \textwidth}
\centering
\includegraphics[width = 1 \textwidth]{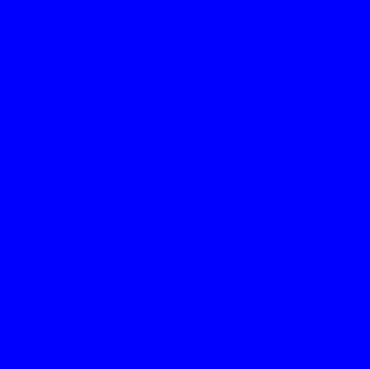}
\label{fig_expl_T1}
\end{minipage}
}
\subfigure[Expl. T2]{
\begin{minipage}{0.3 \textwidth}
\centering
\includegraphics[width = 1 \textwidth]{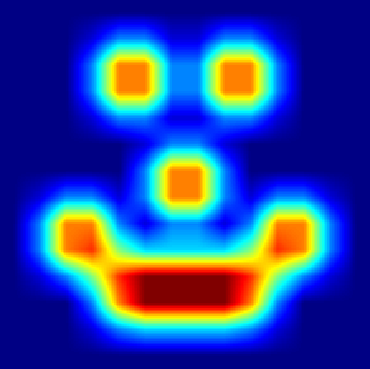}
\label{fig_expl_T2}
\end{minipage}
}
\subfigure[Expl. T3]{
\begin{minipage}{0.3 \textwidth}
\centering
\includegraphics[width = 1 \textwidth]{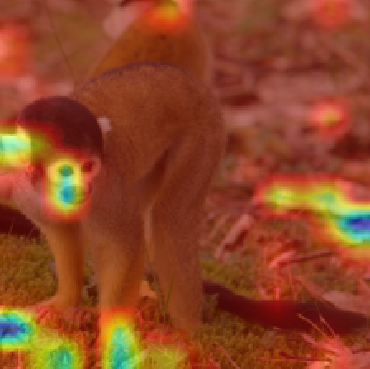}
\label{fig_expl_T3}
\end{minipage}
}
\subfigure[Expl. T4]{
\begin{minipage}{0.3 \textwidth}
\centering
\includegraphics[width = 1 \textwidth]{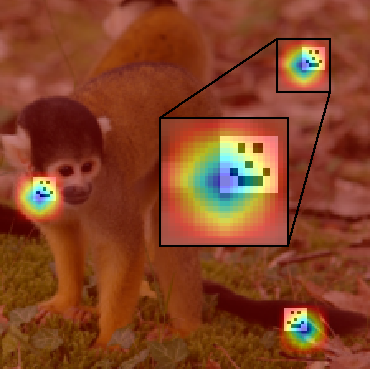}
\label{fig_expl_T4}
\end{minipage}
}
\caption{Qualitative example of manipulated explanations for manipulated networks T1-T4. Blue means a pixel had a large influence on the decision. (c,d) The networks T1 and T2 generate always the same explanation, irrespective of the input to the network. (e) T3 generates a semi-random explanation based on the input. (f) T4 only generates a malicious explanation if a specific pattern (in this case, a smiley) is visible in the input. The area in the square for is enlarged for clarity.}
\label{fig_T123}
\vskip -0.2in
\end{figure}

Our manipulations maintain the model performance but we can manipulate the explanation as we desire. An overview of our proposed techniques T1-T4 are shown in Figure \ref{fig_T123}. 
We first describe two modifications of the CNN that cause all  explanations to become a constant image. Arguably, this manipulation is easy to detect by inspecting the explanations, which is not as easy for the two more techniques that we propose. In one of our techniques the explanation is semi-random and depends on the input. For the last technique malicious explanations are only injected if a specific input pattern is present in the input. These last two techniques are much more difficult to detect using visual inspection of explanations and therefore pose a more serious security concern.

Several works use explanations to localize objects in images \citep{simplegrad,GradCAM,amogh_localization}, which could be used by secondary systems; for example, as a pedestrian detector for a self-driving car or an explanations used by a doctor to find a tumor. Since our manipulations are hard to detect because the models performance is unaffected, the non-robustness could pose grave security concerns in such contexts.

Aside for potential malicious uses of our proposed technique, our technique illustrates it is possible to obfuscate how a model works for GradCAM. Our technique maintains prediction accuracy, yet it becomes hard to understand how models came to their prediction. Thus the model becomes impossible to interpret, while staying useful. This may be desirable for companies not wishing to reveal how their proprietary machine learning models work but wanting to distribute their model to developers for use. Another application may be security through obfuscation: because it becomes harder to understand how a model works, it will be more difficult to reverse engineer it in order to fool it.

\section{GradCAM and Notation}

We briefly review the notation and the GradCAM method \citep{GradCAM}. We only consider CNNs for classification tasks. Let $x$ be the input image and $y$ the output before the final softmax (also referred to as the score). Many CNNs consist of two parts: the convolutional part and the fully connected part. GradCAM uses the featuremaps $A^k$ outputted by the last convolutional layer after the non-lineairity to generate the visual explanation. Here $k = 1,\ldots,K$ indicates the channel, and a single $A^k$ can be regarded as a 2D image. The visual explanation or heatmap $I^c$ for a class $c$ is computed by
\begin{equation}
I^c = \text{ReLU}\left(\sum_k \alpha_k^c A^k\right). \label{eq_I}
\end{equation}
Thus a linear combination of the featuremaps is used to generate the explanation, while the ReLU is used to remove negative values. $\alpha_k^c$ is obtained by global-average-pooling the gradient for class $c$ with respect to the $k$th featuremap,
\begin{equation}
\alpha_k^c = \frac{1}{N_A} \sum_i \sum_j \frac{\partial y^c}{\partial A^k_{ij}}, \label{eq_alpha}
\end{equation}
where $i$ and $j$ are the indices for the pixels in the featuremap  and $N_A$ is the total amount of pixels in the featuremap. Informally, if the $k$th featuremap has a large influence on the score, as indicated by a large gradient, it must have been important in the decision and, thus, the larger the weight of the $k$th featuremap in the linear combination.  

\section{Manipulating the CNN}

We will show several techniques that manipulate the architecture and weights to change the explanation of GradCAM, while keeping the performance of the CNN (more or less) unchanged. The recipe for all these approaches will be the same. Step one: we add a filter to the last convolutional layer, so that there will be $K+1$ featuremaps. The $(K+1)$th featuremap will contain our desired target explanation $I_T$. We will scale $A^{K+1}$ in such a way that $A^{K+1}_{ij} \gg A^k_{ij}$ for all pixel locations $i$, $j$ and channels $k$.
Step two: we change the architecture or weights of the fully connected part, to ensure $\alpha_{K+1}^c \gg \alpha_k^c$ for all $c$ and $k$. Under these conditions, following Equation \ref{eq_I} and \ref{eq_alpha}, the GradCAM explanation will be more or less equal to our desired target explanation, $I^c \approx I_T$ for all $c$. Figure \ref{fig_T123} gives an overview of the techniques T1-T4 which we will now discuss in more detail. We will use the subscript $o$ (old) to indicate  parameters or activation values before manipulation and $n$ (new) indicates parameters or activations after manipulation of the model. 

\subsection{Technique 1: Constant Flat Explanation}

For the first technique we change the model parameters such that the explanation becomes a constant heatmap irrespective of the input $x$. Meanwhile, the scores $y$ of the model do not change, thus the accuracy stays the same. 

We manipulate the network as follows.  
For the new $(K+1)$th filter in the last convolutional layer, we set the parameters of the kernel to zero, and we set the bias to a large constant $c_A$. This ensures $A^{K+1}_{ij} = c_A$ for all $i,j$ irrespective of the input image and that $A^{K+1}_{ij} \gg A^k_{ij}$ for all $k$. Let $Z$ be the last featuremap in the convolutional part of the model. Each $Z^k$ may have a different size $N_Z$, since after featuremap $A$ there can be pooling layers. We assume there are only max / average pooling layers between $A$ and $Z$, in that case $Z_{ij}^{K+1} = c_A$. Let $z$ be the vector obtained by flattening the last featuremaps $Z^k$. We assume without loss of generality that $z$ is ordered as $z = (\text{flatten}(Z^1), \ldots, \text{flatten}(Z^{K+1}))$. Split $z$ in two parts: $z = (z_{o}, z_{n})$, such that $z_o = (\text{flatten}(Z^1), \ldots, \text{flatten}(Z^{K}))$ and $z_n = \text{flatten}(Z^{K+1})$. Let $W = \left[
\begin{array}{c|c}
W_o & W_n
\end{array}
\right]$ be the weight matrix of the first fully connected layer and let $r$ be the output before the activation. $$r_o = W_o z_o + b_o,$$ where $b_o$ is the old learnt bias. For the manipulated model $$r_n = W_o z_o + W_n z_n + b_n.$$ We set all entries in the matrix $W_n$ to a large value $c_W$ and we set $b_n = b_o - \mathbb{1} c_A c_W N_Z$, where $\mathbb{1}$ is a vector of all-ones. Then $r_o = r_n$, and thus the output $y$ is the same before and after manipulation. Because $W_n$ is large, small changes in $Z^{K+1}$ lead to large changes in $y$, thus $\alpha_{K+1}^c$ is large. This ensures $\alpha_{K+1}^c \gg \alpha_k^c$. Recall that however, $Z^{K+1}$ is constant.

\subsection{Technique 2: Constant Image Explanation}

In the last technique, the target explanation $I_T$ was a constant. Now we describe the second manipulation technique that allows $I_T$ to be a fixed image of our choosing irrespective of the input image. We use the same technique as before, with two differences. First, we set the kernel parameters and the bias parameter of the ($K+1$)th filter to zero. Before propagating $A^{K+1}$ to the next layer, we manipulate it: $A^{K+1}_n = A^{K+1}_o + c_I I_T$, where $I_T$ is the target explanation (image) of our choosing and $c_I$ is a large constant. This can be seen as a architectural change. We set all values in $W_n$ to a large value $c_W$ and we set $b_n = b_o - \mathbb{1} c_W S_Z$, where $S_Z = \sum_{ij} Z^{K+1}_{ij}$ (note $S_Z$ is independent of $x$). Then again $r_o = r_n$, and thus $y_o = y_n$. The arguments of the previous technique still hold and thus we have $A^{K+1}_{ij} \gg A^k_{ij}$ and $\alpha_{K+1}^c \gg \alpha_k^c$.

\begin{figure}[htb]
    \centering
    \includegraphics[width = 0.95 \textwidth]{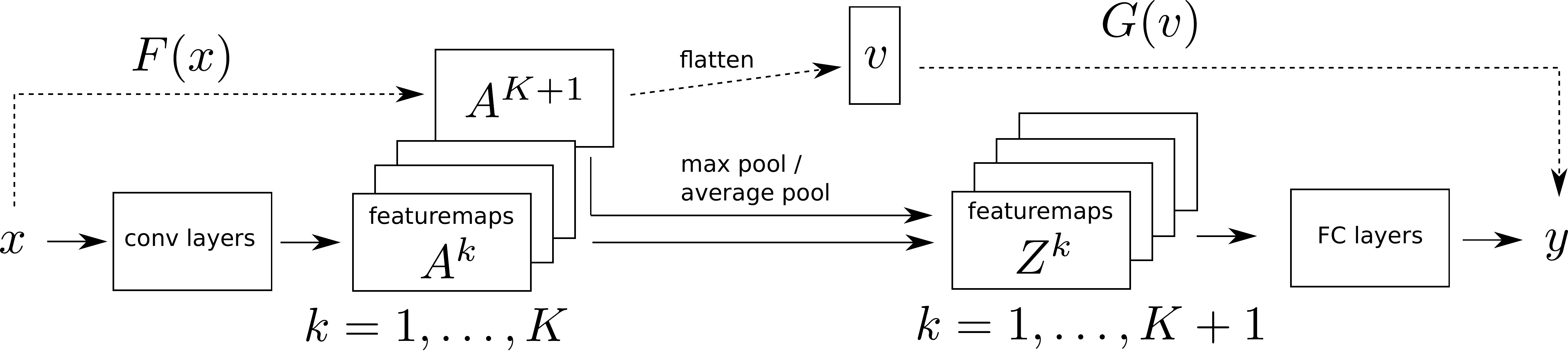}
    \caption{Illustration of architectural changes necessary for techniques T3 and T4. Dashed lines indicate modifications. `conv layers' indicates the convolutional part of the CNN, and the `FC layers' indicate the fully-connected part of the CNN.}
    \label{fig_T34_architecture}
    \vskip -0.3in
\end{figure}

\subsection{Technique 3: Semi-Random Explanation}

A limitation of the previous techniques is that the explanation is always the same irrespective of the input. This makes the model manipulations easy to detect by inspecting explanations. Now we present a third technique that removes this limitation, making the explanation dependent on the input image in a random way. Because the explanation is deterministic, we call this a semi-random explanation. Making the explanation dependent on the input however comes with a price: the scores $y$ may change a small amount of $\epsilon$ and more architectural changes to the model are required. The architectural changes are illustrated in Figure \ref{fig_T34_architecture}.

As before we will put our target explanation $I_T$ in $A^{K+1}$. Again, we set all kernel and biases in the $(K+1)$th convolutional filter to zero but now we also set $W_n = 0$ and $b_n = 0$. To put the target explanation in $A^{K+1}$, we set $A^{K+1}_o = A^{K+1}_n + c_F F(x)$, where $F(x)$ will be a neural network taking $x$ as input and outputs our desired target explanation $I_T$. This can be seen as an architectural change in the form of a branch. We take $F(x)$ to be a randomly initialized CNN (only the convolutional part). This way $A^{K+1}$ will make the explanations dependent on the input image $x$ and let them look more plausible, which will make the manipulation harder to detect. 

To ensure large $\alpha_{K+1}^c$, we add a branch from $A^{K+1}$ to $y$. $\mathbb{1}$ is a vector of all ones. We set $$y_n = y_o + \mathbb{1} G(\text{flatten}(A^{K+1}_n)).$$ $G(v)$ is a scalar valued function taking a vector of length $N_A$ as input. We choose $$G(v) = \epsilon~ \text{mod}(c_G \sum_i v_i, 1),$$ where $mod(a,b)$ is the modulus operator ensures that $G(v) \leq \epsilon$ for all $v$. By choosing $\epsilon$ to be small, the difference between the scores will be small: $|y_n - y_o| \leq \epsilon$.
Furthermore, for all inputs $x$ we have $\frac{\partial G(x)}{\partial x} = \mathbb{1} c_G \epsilon$. By choosing $c_G \gg \epsilon$, we can make the gradient as large as desired, ensuring $\alpha^{K+1}_c$ will be large for all classes $c$. 

\begin{figure}[htb]
\centering

\subfigure[Input (no stickers)]{
\begin{minipage}{0.3 \textwidth}
\centering
\includegraphics[width = 1 \textwidth]{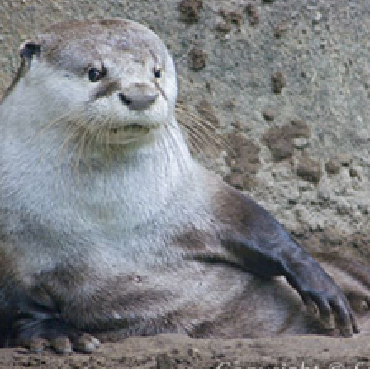}
\label{T4_original}
\end{minipage}
}
\subfigure[Original network (no stickers)]{
\begin{minipage}{0.3 \textwidth}
\centering
\includegraphics[width = 1 \textwidth]{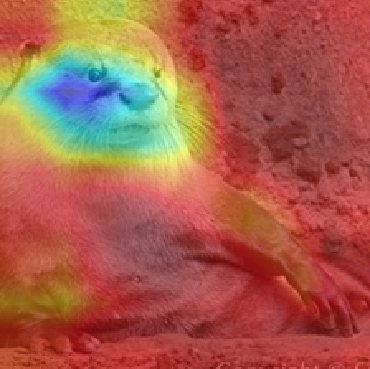}
\label{T4_expl_orig}
\end{minipage}
}
\subfigure[T4 (no stickers)]{
\begin{minipage}{0.3 \textwidth}
\centering
\includegraphics[width = 1 \textwidth]{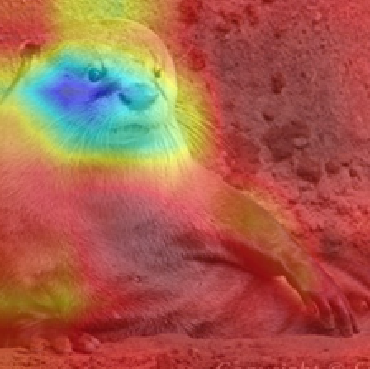}
\label{T4_expl_orig_T4}
\end{minipage}
}
\subfigure[Input (stickers)]{
\begin{minipage}{0.3 \textwidth}
\centering
\includegraphics[width = 1 \textwidth] {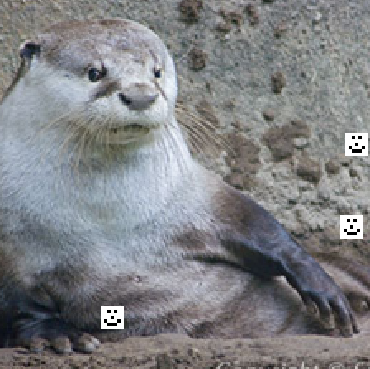}
\label{T4_orig_sticker}
\end{minipage}
}
\subfigure[Original network (stickers)]{
\begin{minipage}{0.3 \textwidth}
\centering
\includegraphics[width = 1 \textwidth]{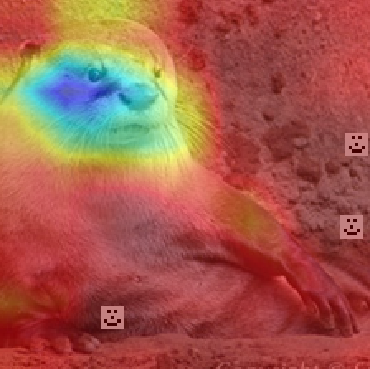}
\label{T4_expl_sticker}
\end{minipage}
}
\subfigure[T4 (stickers) ]{
\begin{minipage}{0.3 \textwidth}
\centering
\includegraphics[width = 1 \textwidth]{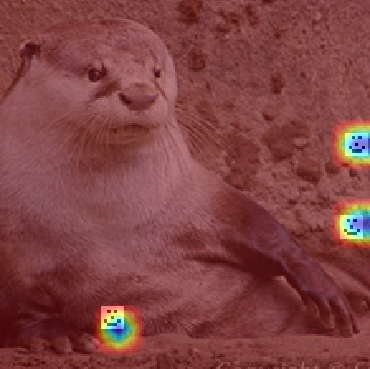}
\label{T4_expl_sticker_T4}
\end{minipage}
}
\vskip 0.01in
\caption{Illustration of Technique 4. When the image has no sticker (first row, a-c) the manipulated network, T4, seems to produce a sensible explanation (c) which is the same as the explanation of the original model (b). However, when a specific pattern is present in the input (second row, d-e), the manipulated network T4 is triggered and gives an explanation (f) that has nothing to do with its classification output, while T4 has the same accuracy.}
\label{fig_T4_example}
\vskip -0.2in
\end{figure}

\subsection{Technique 4: Malicious Explanation Triggered by Input Pattern}

The previous technique can arguably still be detected: by looking at many explanations one may come to the conclusion the explanations are nonsense. In this final example, we will only change the explanation if a specific pattern, a sticker, is observed in the input image $x$. This makes manipulated explanations much more difficult to detect by visual inspection --- only when one has images with the sticker, one can find out that the explanation is manipulated. A visual example is given in Figure \ref{fig_T4_example}. 

We use exactly the same setup as in Technique 3, except that we change $F(x)$. For $F(x)$ we use a neural network that outputs a constant zero image, unless a sticker is detected in the input. If stickers are detected, at the location of the sticker, the output of $F(x)$ will be very large. Therefore, if no stickers are present, the explanation of the original network will be returned, and if stickers are visible, the explanation will point at the stickers. Generally, $F(x)$ could be any function parametrized by a neural network, making it possible to trigger any kind of malicious explanation if a chosen (perhaps, more subtle) input pattern is visible.

\section{Experimental Setup}

For all experiments, we use the VGG-16 network \citep{Simonyan15}. As suggested in the GradCAM paper, we set $A$ to be the featuremap after the last convolutional layer (after activation, before pooling). For VGG-16, $K = 512$ and the resolution of $A^k$ is $14 \times 14$. We evaluate the original network and manipulated networks on the validation set of Imagenet of the ILSVRC2012 competition  \citep{russakovsky2015imagenet}. We generate the heatmap for the class with the highest posterior. The heatmap $I^c$ always has positive values due to the ReLU operation. We normalize all heatmaps by the largest value in the heatmap to map it to $[0,1]$: $\tilde{I}^c = \frac{I^c}{\max_{i,j} I^c_{ij}}$. We measure to what extent our manipulations are successful by measuring the distance between our target explanation $\tilde{I}_T$ and manipulated explanation $\tilde{I}_n$ in terms of the $L_1$ distance. For the experiments with the network T4, we evaluate on the original Imagenet validation set and the manipulated validation set. Manipulated images have 3 randomly placed smiley patterns.

For T1, set $c_A = 100$, $c_W = 100$. For T2, set $c_W = 10$ and we set $I_T$ to a $14 \times 14$ smiley image. For T3, choose $\epsilon = 0.01$, $c_G = 10000$ and $c_F = 1E7$. The network $F(x)$ has a conv2d layer with 6 filters, with filtersize $6 \times 6$, with 3 pixels zero padding at each side, with ReLU activation, followed by a second conv2d layer with 1 filter, kernel size $6 \times 6$, 3 pixels zero padding at each side, with ReLU activation. All weights are randomly initialized. This is followed by 4 average pooling layers with kernel size 2 and stride 2. Then the output of $F(x)$ is $14 \times 14$ and, thus, matches the size of $A^{K+1}$ for VGG-16. For T4 we use a network $F(x)$ that has only one conv2d layer. The smiley pattern is binary: each pixel is white or black. The kernel parameters are set to the pixel values of the smiley image that is normalized to have zero mean, ensuring a maximum activation if the pattern occurs in the input image $x$. We set the bias of the convolutional layer to $b = -\sum_{ij} I_{ij}^2 (1-\frac{1}{N} \sum_{ij} I_{ij}) + 0.0001$ where $I_{ij}$ are the pixel values of the non-normalized smiley image. If the pattern is detected the output is $0.0001$, typically otherwise the output will be negative. We use a ReLU to suppress false detections, followed by 4 average pool layers with same size and stride as before, in order to get the output of $F(x)$ the size $14 \times 14$ and we set $c_F = 1E9$.

\section{Results}

The results for techniques T1-T3 are shown in Table~\ref{t_res1}, for qualitative results see Figure~\ref{fig_T123}. A minimal change in accuracy and scores is observed. After thorough investigation, we found that the change in score and accuracy for T1 and T2 is caused by rounding errors due to the limited precision used in our PyTorch implementation that uses \verb-float16- values --- theoretically, the networks should output the exact same scores and thus the accuracy should stay exactly the same. The $L_1$ distance between our desired target explanation and our observed manipulated explanation is quite small, which matches with the qualitative observation in Figure \ref{fig_T123}. Note that the change in score for T3 is lower than $\epsilon$, as guaranteed.

The results for technique T4 are shown in Table~\ref{t_res2}, for a  qualitative example see Figure~\ref{fig_T4_example}. We observe a small drop in accuracy when the data is manipulated by stickers, as expected, but the accuracy for T4 and the original network are exactly the same. The change in score is very small. If there are no stickers, the target explanation $\tilde{I}_T$ is equal to the explanation of the original network. If there are stickers, $\tilde{I}_T$ is equal to the heatmap that detects the stickers. The observed explanation when a sticker is present is almost equal to the target explanation. Just as desired, if no sticker is present, the explanation of T4 remains the same as the explanation of the original network.

\begin{table}[tbh]
\begin{center}
\vskip -0.1in
\begin{tabular}{@{}llll@{}}
\toprule
                 & Accuracy & $||y_o-y_n||_\infty$ & $||\tilde{I}_T-\tilde{I}_n||_1$ \\ \midrule
Original network & 0.71592                    & -                           & -                            \\
T1: constant     & 0.71594                    & 0.01713                     & 0.00513                      \\
T2: smiley       & 0.71594                    & 0.00454                     & 0.01079                      \\
T3: random       & 0.71592                    & 0.00000                     & 0.05932                             \\ \bottomrule
\end{tabular}
\end{center}
\vskip -0.1in
\caption{Evaluation of manipulated networks T1-T3 on the ILSVRC2012 validation set. Observe that the accuracy more or less stays the same. We measure the difference between the score $y_o$ of the original network and new manipulated score $y_n$ (the score is the output before softmax). The difference between the desired target explanation $\tilde{I}_T$ and the actual observed explanation $\tilde{I}_n$ is measured using the $L_1$ distance. The score changes very little while we can  accurately manipulate the explanation as indicated by small $L_1$ distance. }\label{t_res1}
\vskip -0.2in
\end{table}

\begin{table}[tbh]
\begin{center}
\begin{tabular}{@{}lllll@{}}
\toprule
Dataset                       & Network         & Accuracy     & $||y_o-y_n||_\infty$ & $||\tilde{I}_T-\tilde{I}_n||_1$ \\ \midrule
\multirow{2}{*}{Original}   & Original & 0.71592 & -               & -                 \\
                              & T4: backdoor       & 0.71592 & 0.00000         & 0.00000           \\ \midrule
\multirow{2}{*}{Manipulated (sticker)} & Original & 0.69048 & -               & -                 \\
                              & T4: backdoor       & 0.69048 & 0.00000         & 0.00006          \\ \bottomrule 
\end{tabular}
\end{center}
\vskip -0.1in
\caption{Evaluation of Technique 4 on the ILSVRC2012 validation set. Observe that T4 has the same accuracy and scores as the original network for both kinds of data. When presented with input data without stickers, the manipulated network T4 produces the same explanation as the original network. When presented with manipulated data, the manipulated explanation, $\tilde{I}_n$, is almost equal to the desired explanation, $\tilde{I}_T$.}\label{t_res2}
\vskip -0.2in
\end{table}

\section{Discussion}

GradCAM is not `broken' --- for normally trained models, GradCAM has been proven to be useful. GradCAM does not work for adverserially manipulated models such as ours, since it was not designed for that task. However, our models are valid models, with (almost) equal performance. Hence, they should also admit a valid explanation. In fact, in \cite{AxiomaticAttribution} the axiom of Implementation Invariance is defined: two networks that produce the same output for all inputs should admit the same explanation. Clearly, GradCAM does not satisfy this axiom and thus there is room for improvement. One may wonder wether the axiom should be extended to models that return extremely similar predictions, such as T3 and T4. 

Our work reveals that GradCAM relies on unknown assumptions on the network parameters, architecture, etc. It is difficult to rule out that, by accident, a model can be produced, using regular training, where GradCAM explanations may fail. We think it is important to determine what assumptions should be verified for GradCAM to produce accurate explanations, so we can always verify the correctness of GradCAM explanations. 

Our techniques may be extended to fool other explanation methods. Several methods rely on the gradient $\frac{\partial y}{\partial x}$ \citep{AxiomaticAttribution, simplegrad, gradientTimesInput, smoothgrad, LIME}. T3 and T4 show that it is possible to manipulate the gradient, while affecting accuracy only little. So, these methods may also be vulnerable. 

A weakness of our method is that architectural changes are necessary. If the practitioner visualizes the architecture (for example, using TensorBoard in TensorFlow \citep{tensorflow2015-whitepaper}) or inspects the code, he may easily discover that the model has been tampered with. However, we believe similar attacks, where the original architecture is used, should be feasible, which would make the attack much harder to detect. We believe this is possible, since deep networks contain a lot of redundancy in the weights. Weights can be compressed or pruned, freeing up neurons, which then may be used to confuse the explanation. Recently, this area of research has been very active \citep{compressionsurvey1,compressionsurvey2}. For example, \citet{pruning} were able to prune $35\%$ of the weights, while not significantly changing the test accuracy on MNIST. Another approach is Knowledge Distillation (KD), where a larger model (the teacher) can be compressed in a smaller model (the student) \citep{ba2014deep}. Such methods could be combined with our technique to keep the model accuracy more or less the same and to confuse the explanation method, without any architectural changes. We will explore this promising idea in future work. 

\section{Conclusion}

We provided another sanity check in the same vein as \citet{sanity} and we have shown that GradCAM does not satisfy said sanity check. We submit that, for any explanation method, one should consider whether it is possible to change the underlying model such that the predictions change minimally, while explanations change significantly. If this is the case, our work illustrates that the explanation method may be fooled by an attacker with access to the model and the explanations may not be as robust as desired.

\appendix

\bibliography{main}

\end{document}